\documentclass[10pt,twocolumn,letterpaper]{article}

\usepackage{iccv}
\usepackage{times}
\usepackage{epsfig}
\usepackage{graphicx}
\usepackage{amsmath}
\usepackage{amssymb}
\usepackage{mathrsfs}
\usepackage{tabularx}
\usepackage{bbm}
\usepackage{dsfont}
\usepackage{multirow}
\newcommand{\old}[1]{}

\DeclareMathAlphabet\mathbfcal{OMS}{cmsy}{b}{n}
\newcolumntype{P}[1]{>{\centering\arraybackslash}p{#1}}

\usepackage[breaklinks=true,bookmarks=false]{hyperref}

\iccvfinalcopy 

\def\iccvPaperID{****} 

\ificcvfinal\fi

\begin{document}

\title{Context-Aware Image Matting for Simultaneous Foreground and Alpha Estimation}

\author{Qiqi Hou\\
Portland State University\\
{\tt\small qiqi2@pdx.edu}
\and
Feng Liu\\
Portland State University\\
{\tt\small fliu@cs.pdx.edu}
}

\maketitle

\begin{abstract}

Natural image matting is an important problem in computer vision and graphics. It is an ill-posed problem when only an input image is available without any external information. While the recent deep learning approaches have shown promising results, they only estimate the alpha matte. This paper presents a context-aware natural image matting method for simultaneous foreground and alpha matte estimation. Our method employs two encoder networks to extract essential information for matting. Particularly, we use a matting encoder to learn local features and a context encoder to obtain more global context information. We concatenate the outputs from these two encoders and feed them into decoder networks to simultaneously estimate the foreground and alpha matte. To train this whole deep neural network, we employ both the standard Laplacian loss and the feature loss: the former helps to achieve high numerical performance while the latter leads to more perceptually plausible results. We also report several data augmentation strategies that greatly improve the network's generalization performance. Our qualitative and quantitative experiments show that our method enables high-quality matting for a single natural image. Our inference codes and models have been made publicly available at \url{https://github.com/hqqxyy/Context-Aware-Matting}.

\old{Most current deep learning methods only focus on computing the alpha values of unknown pixels. However many applications, such as composing the object into a new scene, requires the foreground color. This paper presents a context-aware matting method for simultaneous foreground color and alpha matte estimation.  Context information makes it easily classify the foreground and background and thus improve the matting result. Our network has two encoders: one for the matting, the other for the context information. The matting encoder will focus on the local features, while the context encoder will extract the global feature, The obtained features are then jointly feed into decoders to predict the final matting result as well as the foreground. We furthermore improve perceptual quality of our results by the perceptual loss. In the last, we proposed several training data augmentation strategies to enhance our model's generalization on the real world examples. Both qualitative and quantitative experiments show that our method provides a practical solution to the real world matting problem.}
\end{abstract}

\vspace{-0.2in}
\section{Introduction}
\label{sec:intro}
\begin{figure*}[ht]
    \small
    \hspace{-0.1in}
	\begin{tabular}{l}
	     \includegraphics[width=1.0\textwidth]{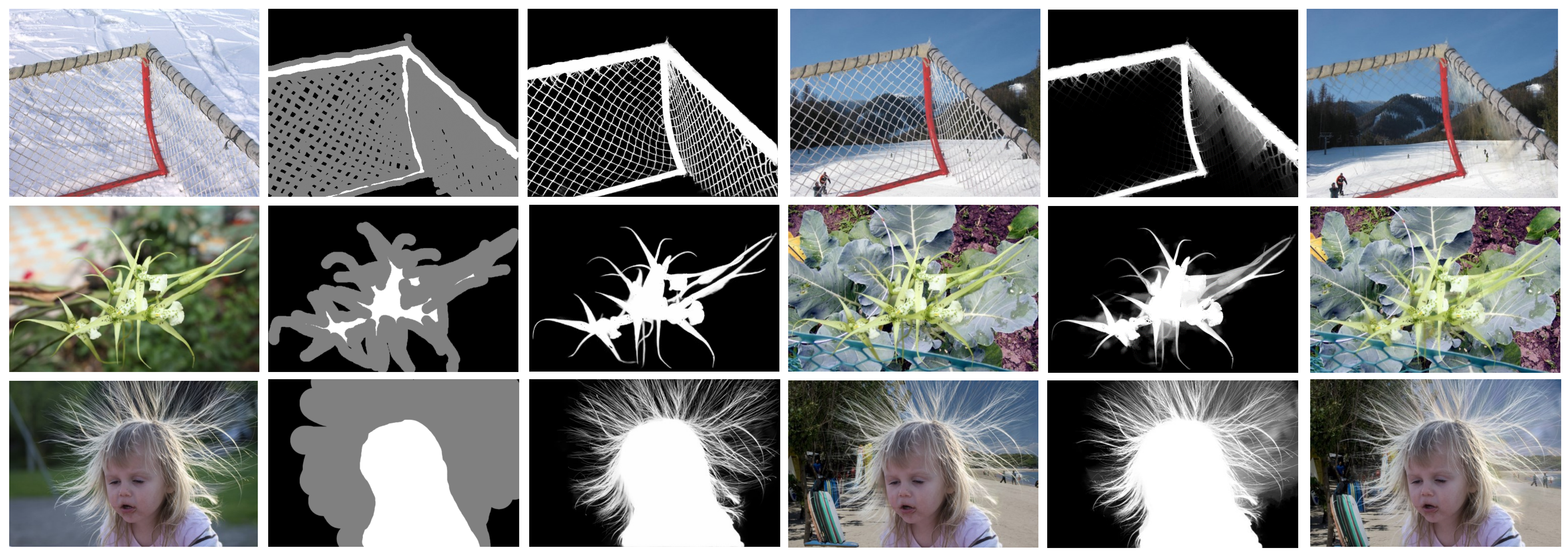}  \\
	     \hspace{0.28in} Input image \hspace{0.6in} Trimap \hspace{0.5in} Our alpha map and composition results \hspace{0.4in} Results from Closed-form~\cite{levin2008closed} 
	\end{tabular}
	
	\caption{Real-world image matting. Our method is able to simultaneously estimate high-quality foreground images and alpha maps from real-world images although trained on a synthetic dataset. Our results keep final structures (the top example) while being free from the common color bleeding problem (the bottom example).  }\vspace{-0.2in}
	\label{fig:motiv}
\end{figure*}

Natural image matting is the problem of estimating the foreground image and the corresponding alpha matte from an input image. It is a critical step of image composition, which is widely used in image and video production. Without any external information, matting is a seriously ill-posed problem. In practice, most existing matting methods take a trimap as input; however, matting is still underconstrained in the undefined area in the trimap.   

Traditional methods solve the matting problem by inferring the alpha matte information in the undefined area from those in the defined areas~\cite{wang2008image}. For instance, the matte values in the undefined areas can be propagated from the known areas according to the spatial and appearance affinity between them~\cite{aksoy2017designing, chen2013knn, chen2013image, he2010fast, lee2011nonlocal, levin2008closed, levin2008spectral,  sun2004poisson}. Alternatively, the undefined matte values can be computed by sampling the color or texture distribution of the known foreground and background areas and optimizing a carefully defined metric, such as the likelihood of the foreground, background, and alpha values~\cite{chuang2001bayesian, he2013iterative, he2011global, wang2005iterative, wang2007optimized}. While these methods provide promising results and some of them are incorporated into commercial tools, single natural image matting is still a challenging problem as these methods rely on the distinctive appearance of the foreground and background areas, such as their local or global color distribution. 

Our research is inspired by the recent deep learning approaches to image matting. These deep matting approaches, such as~\cite{chen2018tom, lutz2018alphagan, xu2017deep} take an input image and the corresponding user-provided trimap as input and output an alpha map. They are shown robust for many challenging scenarios. These methods, however, only output the alpha map without the foreground or background image. 

This paper presents a deep image matting method that simultaneously estimate the alpha map and the foreground image. Our method explores both local image and global context information for high-quality matting. This is inspired by the success of non-deep learning-based matting approaches that combines the global sampling and local propagation strategies~\cite{aksoy2017designing, chen2013knn, chen2013image, he2010fast, he2011global}. Specifically, we designed a two-encoder-two-decoder fully convolutional neural network for context-aware simultaneous foreground image and alpha map estimation. The matting encoder learns to extract the local features while the context encoder learns more global features. We concatenate the features from these two encoders and feed them to an alpha decoder and a foreground decoder to estimate the alpha map and the corresponding foreground image simultaneously. 

We explore a Laplacian loss and the feature loss to train our deep matting neural network. We found that the Laplacian loss enables our network to achieve the state-of-the-art numerical performance while the feature loss leads to more perceptually plausible matting results. We also found that some data augmentation methods are particularly helpful for our neural network to generalize to real-world images although our network is trained on a synthetic dataset provided by Xu~\etal\cite{xu2017deep}.

To our best knowledge, this paper contributes the first deep matting method that enables simultaneous foreground and alpha estimation. Both our qualitative and quantitative experiments demonstrate that our method is able to generate state-of-the-art matting results on challenging real-world examples, as shown in Figure~\ref{fig:motiv}. We attribute the success of our method to 1) the integration of local visual features and global context information, 2) the combination of the Laplacian and feature loss, and 3) various effective data augmentation strategies that help generalizing our method to a wide variety of challenging real-world images.

\section{Related Work}
\label{sec:related} 
\begin{figure*}[ht]
	\centering
	\includegraphics[width=1.0\textwidth]{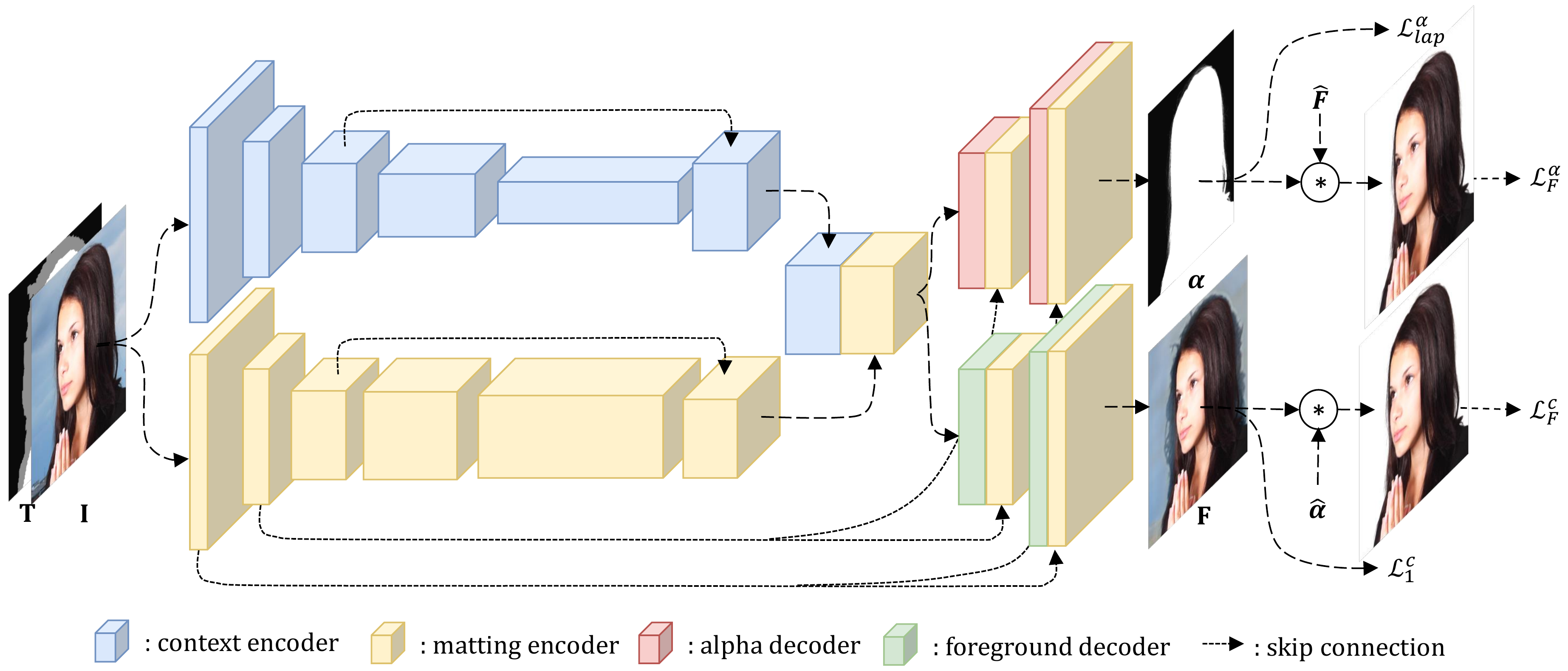}
	\caption{The architecture of our matting network. We design a two-encoder-two-decoder network. The matting encoder and the context encoder capture both visual features and more global context information. The features from these two encoders are concatenated and feed to the foreground and the alpha decoder to output the foreground image and the alpha map of the input image simultaneously.}\vspace{-0.1in}
	\label{fig:arch}
\end{figure*}

Image matting assumes that an image $\mathbf{I}$ is a linear composition of a foreground image $\mathbf{F}$ and a background image $\mathbf{B}$ according to an alpha map $\pmb\alpha$ as follows~\cite{smith1996blue}.
\begin{equation}\label{eq:matt}
	\mathbf{I} = \pmb\alpha \mathbf{F} + (1 - \pmb\alpha) \mathbf{B}
\end{equation}
Given the input image $\mathbf{I}$, image matting aims to recover $\mathbf{F}$, $\mathbf{B}$ and $\pmb\alpha$. Most of existing matting methods require a user-provided trimap that specifies known foreground and background areas, as well as an undefined area. In this way, matting is reduced to solving for the foreground, background, and alpha values in the undefined area. Given only the input image $\mathbf{I}$ and the trimap, matting is a seriously ill-posed problem. A rich literature exists for matting. These methods infer the matte information for the undefined area from the known foreground and background according to the trimap. They either propagate the matte information from the neighboring foreground or background areas to the unknown areas~\cite{aksoy2017designing, chen2013knn, chen2013image, he2010fast, lee2011nonlocal, levin2008closed, levin2008spectral,  sun2004poisson}, or more globally sample the appearance information of the known foreground and background and use them to optimize for the matting in the unknown area~\cite{chuang2001bayesian, he2013iterative, he2011global, wang2005iterative, wang2007optimized}. There are also methods that combine the local propagation strategy and the global sampling strategy to achieve more reliable results~\cite{aksoy2017designing, chen2013knn, chen2013image, he2010fast, he2011global}. Wang and Cohen provided a good survey on these traditional image matting algorithms~\cite{wang2008image}. Our design of a double-encoder-double-decoder network to learn to estimate local and global context information is inspired by these hybrid methods.

Our work is most relevant to the recent deep learning approaches to image matting. Shen~\etal  trained a dedicated deep convolutional neural network for portrait matting~\cite{shen2016deep}. Their method first employs a deep neural network to generate the trimap of a portrait image and then feeds it to an off-the-shelf matting method, namely the Closed-form Matting algorithm~\cite{levin2008closed}, to obtain the final matting result. Cho~\etal developed a deep matting method that takes the matting results from the Closed-form Matting algorithm~\cite{levin2008closed} and the KNN Matting algorithm~\cite{chen2013knn} as input, and refine it using a deep neural network~\cite{cho2016natural, cho2019deep}. Xu~\etal developed a large-scale synthetic image matting dataset and used it to train a two-stage deep neural network for alpha matting. Their method produced high-quality matting results for both synthetic and real-world images~\cite{xu2017deep}. Lutz~\etal explores generative adversarial networks to achieve high-quality natural image matting~\cite{lutz2018alphagan}.   In their recent work, Chen~\etal addressed a difficult case of image matting, transparent object matting. By considering transparent object matting as a refractive flow estimation problem, they developed a two-stage neural network to estimate the refractive flow from only one input image for transparent object matting~\cite{chen2018tom}. While these methods are able to estimate high-quality alpha maps, they do not generate the foreground component. Our work builds upon these deep learning methods and simultaneously estimate the foreground image and the alpha map, thus providing a complete solution to image matting. Our network learns to extract both local visual features and global context information to obtain high-quality image matting.

\section{Context-Aware Image Matting}
\label{sec:method}

Our method takes an image  $\mathbf{I}$ and a user-specified trimap $\mathbf{T}$ as input and aims to estimate the foreground $\mathbf{F}$ and the corresponding alpha map  $\pmb\alpha$, thus providing a full solution to matting. With the foreground and the alpha map, we can directly compute the background according to Equation~\ref{eq:matt}.

We design a context-aware two-encoder-two-decoder deep neural network to simultaneously estimate the foreground and the alpha map, as shown in Figure~\ref{fig:arch}. The outputs of the two encoders are concatenated and fed to the two decoder to generate the foreground and the alpha map, respectively. The two-encoder design of the network is inspired by the success of traditional matting algorithms that combine the local propagation and global sampling strategies for robust image matting~\cite{aksoy2017designing, chen2013knn, he2011global, he2010fast}. Specifically, the matting encoder is designed to learn to extract local features that are required to capture final image structures, such as hairs, while the context encoder learns to estimate more global context information that is helpful to disambiguate the foreground and background when they are similar to each other locally. Below we describe the encoders and decoders in more detail.

\textbf{Matting encoder.}  We adopt the modified version of the Xception 65 architecture~\cite{chollet2017xception} from the deeplab v3+\cite{deeplabv3plus2018} and set the down-sampling factor as 4 by setting the \texttt{entroy flow}'s \texttt{block2} and \texttt{block3}'s stride as 1. This modification enables the \texttt{middle flow} to have a big spatial resolution.  While traditional classification models~\cite{chollet2017xception, he2016deep, krizhevsky2012imagenet, mobilenetv22018, simonyan2014very} more aggressively compromise the spatial resolution to have a large valid receptive field, we use such a smaller down-sampling factor to retain sufficient spatial information that is important for the task of matting to capture fine image structures. Meanwhile, there is a trade-off between the computation/memory cost and spatial resolution. We empirically find that the down-sampling factor of 4 can get good matting results and cost a relatively small amount of computation and memory. We use skip connections to use features from the earlier layers as shown in Figure~\ref{fig:arch}.

\noindent\textbf{Context encoder.} We also adopt the Xception 65 architecture~\cite{chollet2017xception} from \cite{deeplabv3plus2018}. Compared to the matting encoder, we use a much larger down-sampling factor of 16 to capture more global contextual information. We bilinearly upsample the final features by a factor of 4 so that the context features are of the same size as the local matting features from the matting encoder.

\noindent\textbf{Alpha decoder} and \textbf{foreground decoder} have the same network architecture. Specifically, we first bilinearly upsample the concatenated features from the encoders by a factor of 2 and then combine them with the intermediate features from the context encoder using a skip connection as shown in Figure~\ref{fig:arch}. This is followed by two $3 \times 3$ convolutional layers with 64 channels. We repeat this process twice so that each decoder outputs the foreground image and the alpha map with the same size as the input image.

\subsection{Loss functions} 
\label{sec:loss}

We compute the loss over both the alpha map and the foreground image. We explore a range of loss functions to train our network. Below we describe them one by one.

We use a Laplacian loss~\cite{niklaus2018context} to measure the difference between the predicated alpha map $\pmb\alpha$ and its ground truth $\hat{\pmb\alpha}$.
\begin{equation}
\mathcal{L}_{lap}^{\alpha} = \sum_{i = 1}^5 2^{i-1} \Arrowvert L^i(\hat{\pmb\alpha}) - L^i(\pmb\alpha) \Arrowvert_1,
\end{equation}
where $L^i(\pmb\alpha)$ indicates the $i^{th}$ level of the Laplacian pyramid of the alpha map. This loss function measures the differences of two Laplacian pyramid representations and captures the local and global difference. We scale the contribution of a Laplacian level according to its spatial size.

We also use the feature loss to measure the perceptual quality of the alpha map. The feature loss, based on the differences between the high-level features extracted from a pre-trained convolutional neural network, has been shown effective in generating perceptually high-quality images in many image enhancement and synthesis tasks~\cite{dosovitskiy2016generating, ledig2017photo, niklaus2018context, niklaus2017video, sajjadi2017enhancenet,zhang2018unreasonable, zhu2016generative}. However, it is difficult to directly measure the perceptual quality of an alpha map. Our solution is to composite the ground-truth foreground image onto the black background using the alpha map and then measure the perceptual quality of the composition result as follows.
\begin{equation}
\mathcal{L}_{F}^{\alpha} = \sum_{layer} \Arrowvert \phi_{layer}(\hat{\pmb\alpha} * \hat{\textbf{F}}) - \phi_{layer}(\pmb\alpha * \hat{\textbf{F}}) \Arrowvert_2^2,
\end{equation}
where $\hat{\textbf{F}}$ indicates the ground truth foreground and $\phi_{layer}$ indicates the features output by the $layer$ in a pre-trained VGG16 network~\cite{simonyan2014very}. Our method uses  [\texttt{conv1\_2}, \texttt{conv2\_2}, \texttt{conv3\_3}, \texttt{conv4\_3}] to compute the features.

We follow the same setting to calculate the feature loss for the predicated foreground image. Here the feature loss $\mathcal{L}_{F}^{c}$ is computed on the composition result using the ground-truth alpha map with the foreground image as follows.
\begin{equation}
\mathcal{L}_{F}^{c} = \sum_{layer} \Arrowvert \phi_{layer}(\hat{\pmb\alpha} * \hat{\textbf{F}}) - \phi_{layer}(\hat{\pmb\alpha} * \textbf{F}) \Arrowvert_2^2,
\end{equation}

We also use the standard $\ell_1$ loss for the predicted foreground $\textbf{F}$. We only calculate the loss where the foreground is visible, in other words, the ground truth alpha matte is bigger than 0,
\begin{equation}
\mathcal{L}_{1}^{c} = \Arrowvert \mathbbm{1}(\hat{\pmb\alpha} > 0)*(\hat{\textbf{F}} - \textbf{F}) \Arrowvert_1,
\end{equation}
where $\mathbbm{1}$ is an indicator function that takes 1 if the statement is true and 0 otherwise.

Finally, we apply the standard $\ell_2$ regularization loss to all the convolutional layers.  We will examine these loss functions in our experiments (Section~\ref{sec:exp}).

\subsection{Training}
\label{sec:train}

We initialize our neural network with pre-trained models from \cite{deeplabv3plus2018}. We use TensorFlow to train our neural network. Similar to \cite{deeplabv3plus2018}, we use the ``poly" learning rate policy to train our network, where $lr = lr_{init}(1 - \frac{iter}{max\_iter})^{power}$ with $lr_{init} = 7\times 10^{-4}$ and $power = 0.9$. We use a mini-batch size of 6 and train the neural network for 1 million iterations for models (1-3) in Table \ref{table:rst-com1k-matte}. We fine-tune models (4-9) based on the pretrained model (3) with $10^5$ iterations with $lr_{init} = 10^{-4}$. 

\begin{figure}[t]
	\centering
	\includegraphics[width=0.45\textwidth]{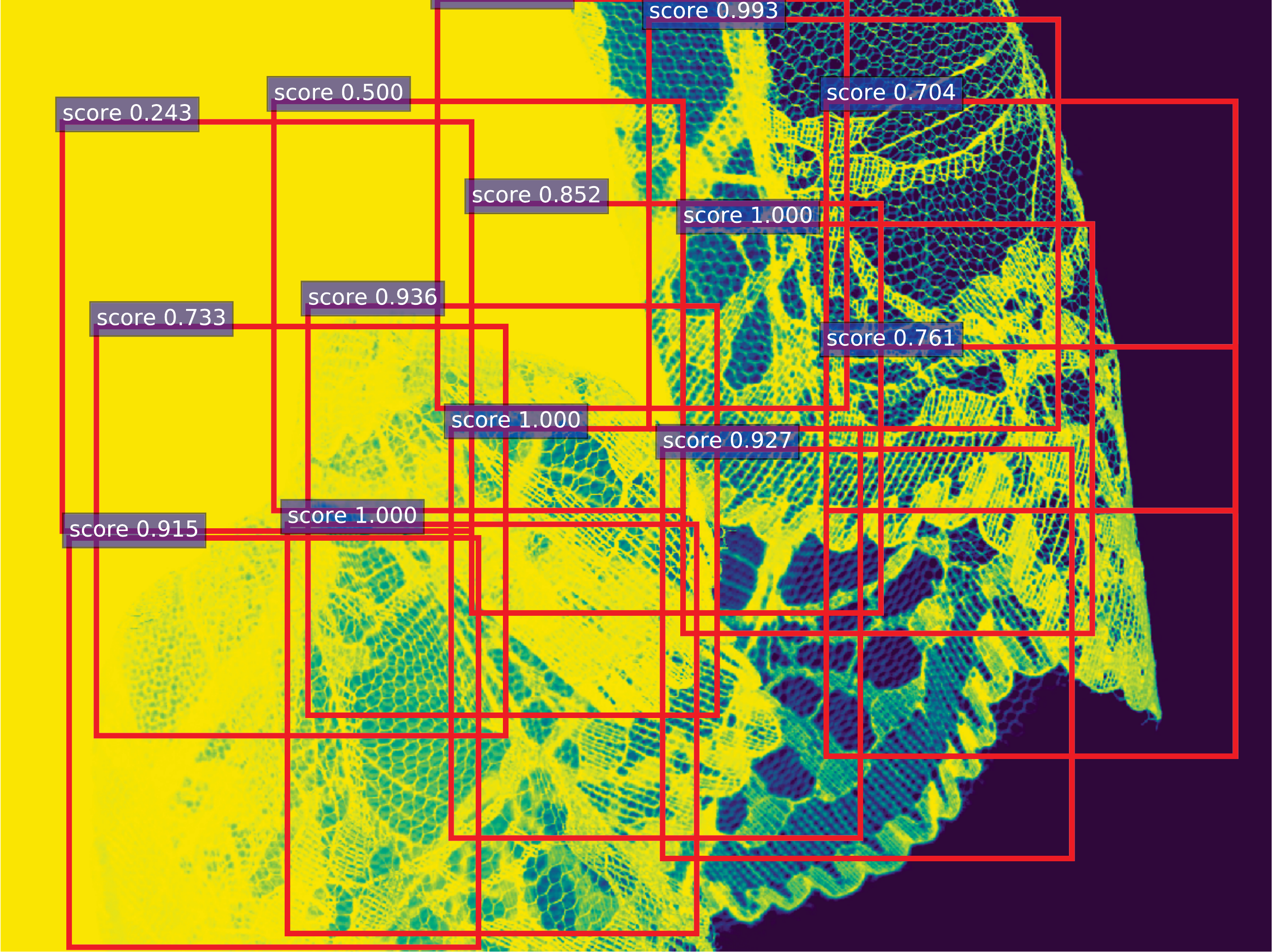}
	\caption{Image patch selection. The alpha map is illustrated using the color map, with yellow and blue indicating the foreground and background, respectively. The patches are selected to cover the unknown region but with relatively small overlaps among them.}\vspace{-0.15in}
	\label{fig:patch-selection}
\end{figure}

\noindent\textbf{Training dataset.} We train our network using the matting dataset shared by Xu \etal\cite{xu2017deep}. This dataset contains 431 training images with the corresponding alpha maps and the foreground images. We create the training samples in a similar way to Xu~\etal. Specifically, we composite the foreground image onto a randomly selected background image from MS-COCO dataset~\cite{lin2014microsoft}. We down-sample the foreground image gradually by a factor of 0.9 until the short side is 600 pixels. If the source image's short side is less than 600 pixels, we first scale it up to 780. In total, we generate 1957 scaled foreground image. Then we select image patches that contain unknown regions in the trimap. Specially, we slide windows of size $600\times600$ on the full image with a stride of 5 pixels to get a large amount of candidate windows and remove patches where less than 10\% pixels are unknown. Furthermore, since many patches overlap with each other significantly, we employ non-maximum suppression(NMS) to remove overlapping patches. Specifically, we set the NMS threshold as 0.3 and only keep the top 30 image patches with the highest unknown pixel percentages in each image. Figure \ref{fig:patch-selection} shows an example of selected image patches. In total, we obtain 9,507 $600 \times 600$ foreground image patches. Finally, we create training samples of size $225 \times 225$ by randomly cropping the composited image with the following data augmentation operators.

\begin{figure}[t]
	\centering
	\includegraphics[width=0.48\textwidth]{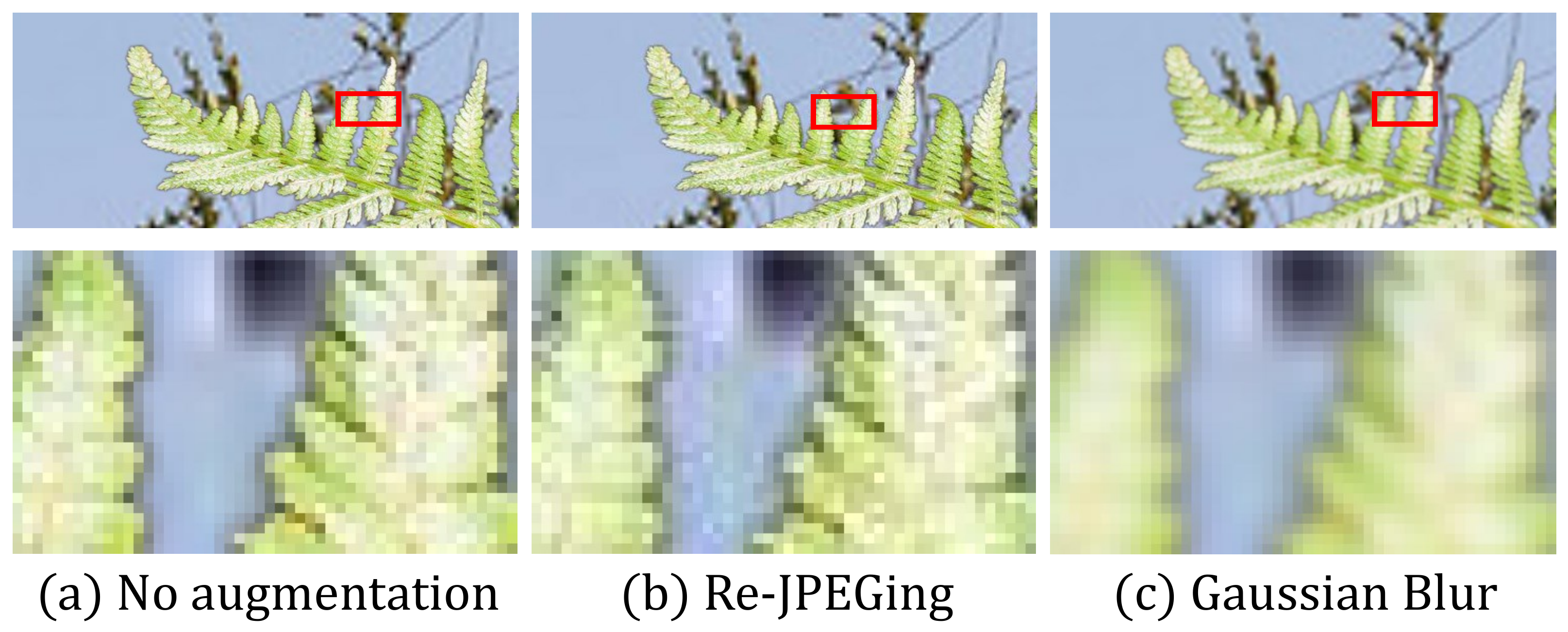}
	\caption{Data augmentation. In the composited image without any data augmentation (a), the foreground image contains some JPEG artifacts while the background is smooth, which produces a bias that will compromises the training of the network. Re-JPEGing introduces the artifacts to the foreground and the background to reduce the bias while Gaussian Blur does so by smoothing the high-frequency artifacts.}\vspace{-0.2in}
	\label{fig:aug-demo}
\end{figure}

\noindent\textbf{Data augmentation.} Following Xu~\etal\cite{xu2017deep}, our training samples are obtained by compositing a foreground image and a background image using an alpha map. As reported in many papers~\cite{agarwal2017photo, de1994learning, ghosh2017detection, huang2010detecting, huh2018fighting,  liu2011detection, luo2010jpeg,  popescu2005exposing, swaminathan2008digital}, many subtle artifacts, such as misaligned JPEG blocks, compression quantization artifacts, and resampling artifacts, can sometimes affect their methods a lot despite that the images look  plausible to the human eyes. Some splice detection methods \cite{agarwal2017photo, huang2010detecting, huh2018fighting,  luo2010jpeg, popescu2005exposing, swaminathan2008digital} even build their algorithms based on such an observation. Directly training the network on the composited images without special augmentation may suffer from a similar problem and thus compromises the generalization capability of the trained network. 

Therefore, besides the resizing augmentation used in Xu \etal\cite{xu2017deep}, we follow the post processing steps in the image splice detection methods~\cite{de2013exposing, huh2018fighting, ng2004data} and use re-JPEGing and Gaussian blur to augment our training samples. These operators  introduce subtle artifacts that are not visually noticeable but can make the network less bias to the small difference between the foreground and the background. As shown in Figure~\ref{fig:aug-demo}, the original background is smoother than the original foreground image. Therefore, it is possible that the network relies on this bias to differentiate the foreground from the background. Re-JPEGing and Gaussian blur can relieve this problem by introducing artifacts or remove these artifacts. For re-JPEGing, we keep 70\% quality of the composited images. For Gaussian blur, we on-the-fly generate a Gaussian kernel with standard deviation in the range of [0, 3] and the kernel size in the range of [3, 5], and apply it to the composited image. We also randomly resize the composited image with a rate of between 0.5 and 1.

Besides, we also use some standard data augmentation operators. Specifically, we employ the gamma transforms to increase the color diversity. The gamma value is randomly selected from [0.2, 2]. We randomly flip the images horizontally. The trimap for each image is automatically generated by randomly dilating its corresponding ground truth alpha map in the range of [4, 25].


\begin{table}[t]
	\centering
	\caption{Alpha map results on the Composition-1K testing set.}
	\footnotesize
	\label{table:rst-com1k-matte}
	\begin{tabular}{lcccc}
			\hline
			\multicolumn{1}{m{35mm}}{Methods} & SAD & MSE($10^3$) & Grad & Conn \\ 
			\hline
			\multicolumn{1}{m{35mm}}{Shared Matting\cite{gastal2010shared}}  & 128.9 & 91 & 126.5 & 135.3 \\
			\multicolumn{1}{m{35mm}}{Learning Based Matting \cite{zheng2009learning}} & 113.9 & 48 & 91.6 & 122.2 \\
			\multicolumn{1}{m{35mm}}{Comprehensive Sampling \cite{shahrian2013improving}} & 143.8 & 71 &102.2 & 142.7 \\
			\multicolumn{1}{m{35mm}}{Global Matting \cite{he2011global}} & 133.6 & 68 & 97.6 & 133.3 \\
			\multicolumn{1}{m{35mm}}{Closed-Form Matting \cite{levin2008closed}} & 168.1 & 91 & 126.9 & 167.9 \\
			\multicolumn{1}{m{35mm}}{KNN Matting \cite{chen2013knn} } & 175.4 & 103 & 124.1 & 176.4 \\
			\multicolumn{1}{m{35mm}}{DCNN Matting \cite{cho2016natural}} & 161.4 & 87 & 115.1 & 161.9 \\
			\multicolumn{1}{m{35mm}}{Three-layer Graph \cite{li2017three}} & 106.4 & 66 & 70.0 & - \\    
			\multicolumn{1}{m{35mm}}{Deep Matting \cite{xu2017deep}} & 50.4 & 14 & 31.0 & 50.8 \\
			\multicolumn{1}{m{35mm}}{Information-flow Matting \cite{aksoy2017designing}} & 75.4 &  66 & 63.0 & - \\
			\multicolumn{1}{m{35mm}}{AlphaGan-Best\footnote{Since AlphaGan only reports the improved results based on other methods, we report the best result of them for every evaluation term} \cite{lutz2018alphagan} }  & 52.4 & 30 & 38.0 & -  \\ 
			\hline
			\multicolumn{1}{m{35mm}}{(1) ME + $\mathcal{L}_{deepmatting}$} & 49.1 &13.4 &26.7  &49.8 \\
			\multicolumn{1}{m{35mm}}{(2) ME + $\mathcal{L}_{lap}^{\alpha}$} &43.9 &11.8 &20.6 &41.6 \\
			\multicolumn{1}{m{35mm}}{(3) ME + CE + $\mathcal{L}_{lap}^{\alpha}$} & \textbf{35.8}  &\textbf{8.2} &17.3  &\textbf{33.2} \\
			\multicolumn{1}{m{35mm}}{(4) ME + CE + $\mathcal{L}_{lap}^{\alpha}$ + $\mathcal{L}_{F}^{\alpha}$} & 38.8  &9.0 &19.0 &36.0 \\
			\multicolumn{1}{m{35mm}}{(5) ME + CE + $\mathcal{L}_{lap}^{\alpha}$ + $\mathcal{L}_{F}^{\alpha}$ + DA} & 71.3  &23.6 &38.8 &72.0 \\
			\multicolumn{1}{m{35mm}}{(6) ME + CE + $\mathcal{L}_{lap}^{\alpha}$ + $\mathcal{L}_{F}^{\alpha}$ +  $\mathcal{L}_{1}^{c}$ +  $\mathcal{L}_{F}^{c}$} & 38.0 & 8.8 & \textbf{16.9} & 35.4 \\
			\multicolumn{1}{m{35mm}}{(7) ME + CE + $\mathcal{L}_{lap}^{\alpha}$ + $\mathcal{L}_{F}^{\alpha}$ +  $\mathcal{L}_{1}^{c}$ +  $\mathcal{L}_{F}^{c}$ + DA} & 84.1 & 29.1 & 39.2 & - \\
			\multicolumn{1}{m{35mm}}{(8) ME + CE + $\mathcal{L}_{lap}^{\alpha}$ + $\mathcal{L}_{F}^{\alpha}$ +  $\mathcal{L}_{1}^{c}$ +  $\mathcal{L}_{F}^{c}$ + DA - ReJPEGing} & 55.1 & 15.5 & 24.6 & 54.7 \\
			\multicolumn{1}{m{35mm}}{(9) ME + CE + $\mathcal{L}_{lap}^{\alpha}$ + $\mathcal{L}_{F}^{\alpha}$ +  $\mathcal{L}_{1}^{c}$ +  $\mathcal{L}_{F}^{c}$ + DA - GaussianBlur} & 69.1 & 23.5 & 39.6 & 69.1 \\
			\hline
	\end{tabular}\vspace{-0.2in}
\end{table}

\section{Experiments}
\label{sec:exp}
We experiment with our methods on the synthetic Composition-1K dataset and a real-world matting image dataset, both of which are provided by Xu~\etal\cite{xu2017deep}. As discussed in Section~\ref{sec:train}, our neural networks are all trained on the synthetic Composition-1K training set. We evaluate our models and compare to the state of the art methods on the Composition-1K testing set and the real-world matting image set. Specifically, the Composition-1K testing dataset contains 1000 composited images. They were generated by compositing 50 unique foreground images onto each of the 20 images from the PASCAL VOC 2012 dataset~\cite{everingham2010pascal}. We used the code provided by Xu~\etal\cite{xu2017deep} to generate these testing images. The real world image dataset contains 31 real world images pulled from the internet~\cite{xu2017deep}. We conduct our user study on the real world images.  

Since not all the methods produce both the foreground images and the alpha maps as the final matting results, we compare our methods to the state of the art on the alpha maps and the foreground images separately. Besides, we also report our user study and our ablation studies to more thoroughly evaluate our methods.

\subsection{Evaluation on alpha maps}
\label{sec:exp:alpha}

We compare our methods to both the state of the art non-deep learning methods, including Shared Matting~\cite{gastal2010shared}, Learning Based Matting~\cite{zheng2009learning}, Comprehensive Sampling~\cite{shahrian2013improving}, Global Matting~\cite{he2011global}, Closed-form Matting~\cite{levin2008closed}, KNN Matting~\cite{chen2013knn}, Three-layer Graph \cite{li2017three}, Information-flow Matting \cite{aksoy2017designing}, and recent deep learning matting approaches, including DCNN Matting \cite{cho2016natural}, Deep Matting \cite{xu2017deep} and AlphaGan~\cite{lutz2018alphagan}. Table~\ref{table:rst-com1k-matte} reports the results on these methods as well as ours on the Composition-1K dataset. The results of the comparing methods are obtained either from their papers or from the recent studies~\cite{lutz2018alphagan,xu2017deep}.

To evaluate these methods, we use various metrics, including SAD, MSE, Gradient (Grad) and Connectivity (Conn)~\cite{rhemann2009perceptually}. Note that the Conn metric fails on some results, which are denoted as ``-". For the ablation analysis of our work, we reported our results on nine versions of our networks with different components. We use ``ME", ``CE", ``DA" to indicate the matting encoder, the context encoder, and data augmentation, respectively. 

\begin{table}[t]
	\centering
	\caption{The foreground result on the Composition-1k dataset.}
	\label{table:rst-com1k-fg}
	\small
	\begin{tabular}{lcc}
			\hline
			\multicolumn{1}{m{50mm}}{Methods} & SAD & MSE($10^3$) \\ 
			\hline

			\multicolumn{1}{m{50mm}}{Global Matting \cite{he2011global}} & 220.39 &36.29  \\
			\multicolumn{1}{m{50mm}}{Closed-Form Matting \cite{levin2008closed}} & 254.15 & 40.89 \\
			\multicolumn{1}{m{50mm}}{KNN Matting \cite{chen2013knn}} & 281.92 & 36.29 \\
			\hline
			\multicolumn{1}{m{50mm}}{(6) ME + CE + $\mathcal{L}_{lap}^{\alpha}$ + $\mathcal{L}_{F}^{\alpha}$ +  $\mathcal{L}_{1}^{c}$ +  $\mathcal{L}_{F}^{c}$} & 61.72 & 3.24  \\
			\multicolumn{1}{m{50mm}}{(7) ME + CE + $\mathcal{L}_{lap}^{\alpha}$ + $\mathcal{L}_{F}^{\alpha}$ +  $\mathcal{L}_{1}^{c}$ +  $\mathcal{L}_{F}^{c}$ + DA} & 94.41 & 8.67  \\
			\multicolumn{1}{m{50mm}}{(8) ME + CE + $\mathcal{L}_{lap}^{\alpha}$ + $\mathcal{L}_{F}^{\alpha}$ +  $\mathcal{L}_{1}^{c}$ +  $\mathcal{L}_{F}^{c}$ + DA - ReJPEGing} &73.79 & 4.96  \\
			\multicolumn{1}{m{50mm}}{(9) ME + CE + $\mathcal{L}_{lap}^{\alpha}$ + $\mathcal{L}_{F}^{\alpha}$ +  $\mathcal{L}_{1}^{c}$ +  $\mathcal{L}_{F}^{c}$ + DA - GaussianBlur} & 85.8 &7.10  \\
			\hline
	\end{tabular}\vspace{-0.1in}
\end{table}

\begin{table}[t]
\setlength{\tabcolsep}{3pt}
	\caption{ Parameter numbers of our models and their performance on the Composition-1K dataset.}
	\footnotesize
	\label{table:rst-param-com1k-matte}
	\begin{tabular}{lcccccc}
			\hline
			\multicolumn{1}{m{25mm}}{Methods}& \# of Parameters & SAD & MSE($10^3$) &Grad & Conn \\ 
			\hline
			\multicolumn{1}{m{25mm}}{ME (model 2)} & 54.0 M &43.9 &11.8 &20.6 &41.6 \\
			\multicolumn{1}{m{25mm}}{ME (deeper model 2)}& 117.0 M & 43.7  &11.0 &21.2    &42.6 \\
			\multicolumn{1}{m{25mm}}{ME + CE (model 3) }& 107.5 M & \textbf{35.8}  &\textbf{8.2} &\textbf{17.3}  &\textbf{33.2} \\
			\hline
	\end{tabular}
\end{table}

\begin{table}[t]
	\centering
	\caption{Comparison of visual quality on the real-world dataset.}
	\label{tbl:nima}
	\begin{tabular}{lcc}
			\hline
			\multicolumn{1}{m{40mm}}{Methods} &Mean score &Std \\ 
			\hline
			\multicolumn{1}{m{40mm}}{ME + CE + $\mathcal{L}_{lap}$} &4.64 &0.42\\
			\multicolumn{1}{m{40mm}}{ME + CE + $\mathcal{L}_{lap}$ + $\mathcal{L}_{F}$ } &4.69 &0.40\\
			\multicolumn{1}{m{40mm}}{ME + CE + $\mathcal{L}_{lap}$ + $\mathcal{L}_{F}$ + DA} &5.03 &0.25\\
			\hline
	\end{tabular}\vspace{-0.25in}
\end{table}


As shown in Table~\ref{table:rst-com1k-matte}, our two-encoder-two-decoder models (model (3-9)) generate matting results with significantly smaller errors than the state of the art methods. To understand what contributes to this improvement, we evaluated on a baseline method (model (2)) that removes the context encoder and found that this baseline model performs much worse according to all the four metrics. Therefore, the improvement can be mainly attribute to the use of our two encoders to capture both local visual features for fine structures and more global contextual information to disambiguate the locally similar foreground and background. Besides these numerical scores, our methods produce visually more plausible results as shown in Figure~\ref{fig:real-mat}. For example, the last example has a strand of long hair. The results from existing methods either miss it entirely or the hair is broken into pieces while our methods better preserve it.

\noindent\textbf{Number of parameters.} We make model (2) deeper so that its number of parameters roughly match model (3). As shown in Table~\ref{table:rst-param-com1k-matte}, while this deeper version of model (2) improves over the original one w.r.t SAD and MSE, it  performs worse than our model (3) (ME + CE).

\begin{figure}[t]
	\centering
	\includegraphics[width=0.5\textwidth]{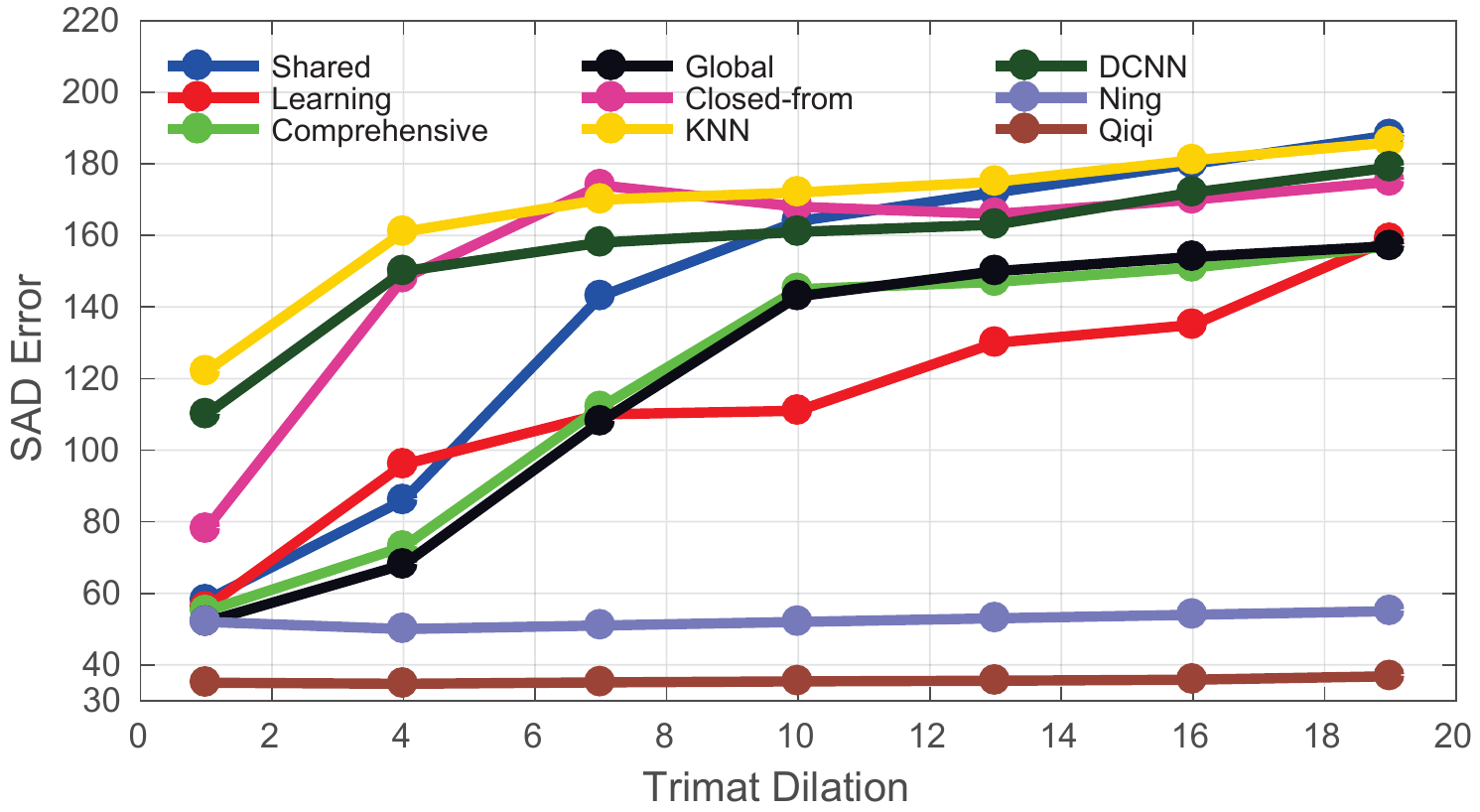}\vspace{-0.05in}
	\caption{Sensitivity test with respect to trimap sizes.}\vspace{-0.2in}
	\label{fig:ana-trimap} 
\end{figure}

\begin{figure*}[t]
	\centering
	\includegraphics[width=1.0\textwidth]{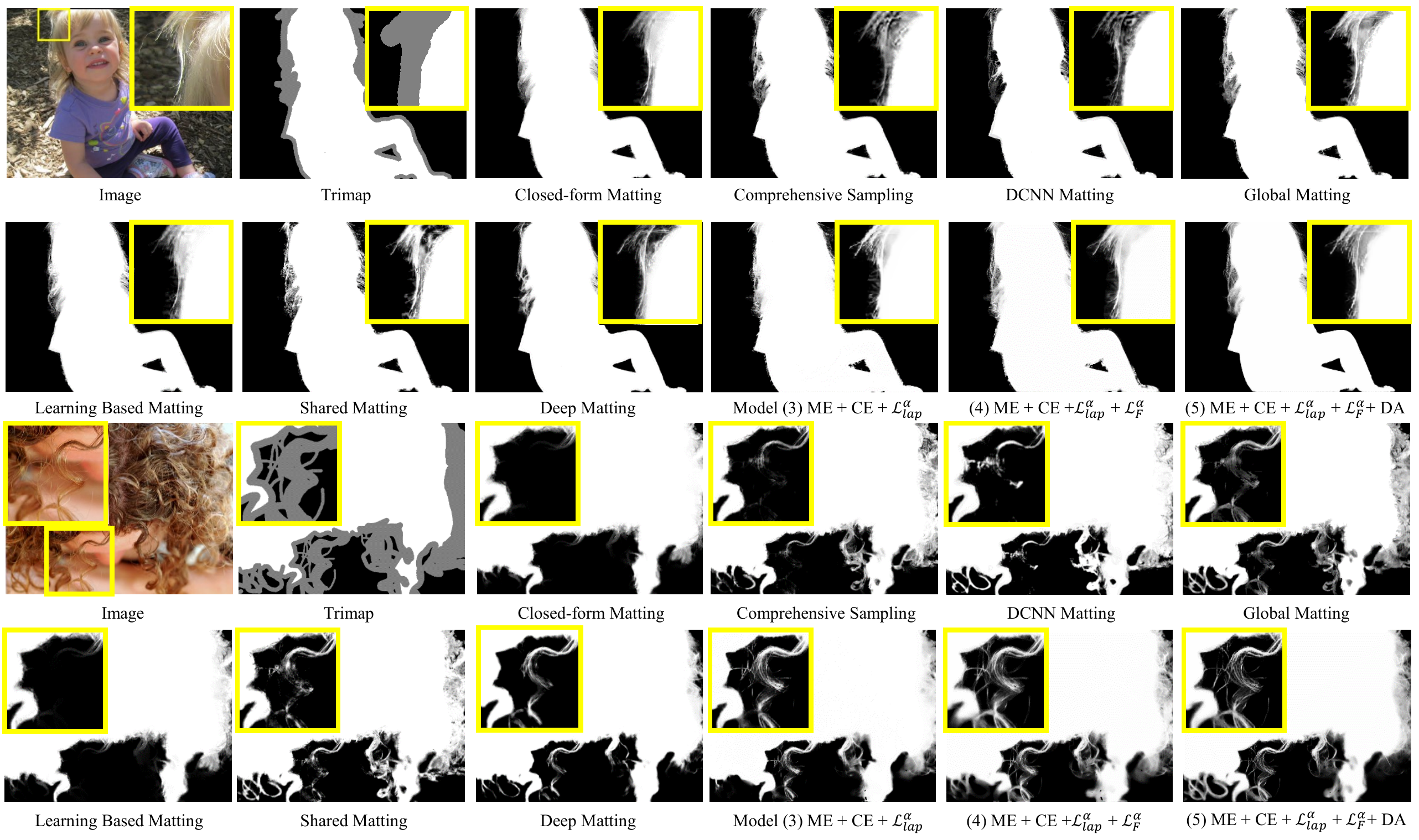}\vspace{-0.06in}
	\caption{Comparison of the alpha matte on the real world images dataset \cite{xu2017deep}.}\vspace{-0.225in}
	\label{fig:real-mat}
\end{figure*}

\noindent\textbf{Sensitivity of trimap.} Following the same process of the Deep Matting work~\cite{xu2017deep}, we examine our method's sensitivity to the trimap size by dilating the ground-truth to a range of sizes. As illustrated in Figure~\ref{fig:ana-trimap}, our method is stable to the trimap sizes. Note, the scores of comparing methods were obtained from~\cite{xu2017deep}.

\subsection{Evaluation on foreground images}

As existing deep learning methods only output alpha maps,  we compare to three representative non-deep learning matting methods, namely Global Matting~\cite{he2011global}, Closed-Form Matting~\cite{levin2008closed} and KNN Matting~\cite{chen2013knn}, on how well foreground images can be extracted from single input images on the Composition-1K dataset.  We calculate the SAD and MSE of $\pmb\alpha * \mathbf{F}$ following the previous work~\cite{price2010simultaneous}. As shown in Table~\ref{table:rst-com1k-fg}, our method reduces the error by a large margin.

\subsection{Ablation study}

As discussed in Section~\ref{sec:exp:alpha}, our two-encoder structure brings in the major performance improvement. Besides, we found that proper loss functions and data augmentations are also important to obtain high-quality matting results and help generalizing to real-world images.

\noindent\textbf{Loss functions.} As shown in Table~\ref{table:rst-com1k-matte}, our model (2) with the Laplacian loss $\mathcal{L}_{lap}^{\alpha}$ generates more numerically accurate results than our model (1) with the loss used in Deep Matting~\cite{xu2017deep}. Our model (3) generates better result compared to the model (4) with both the Laplacian loss and the feature loss $\mathcal{L}_{F}^{\alpha}$. On the other hand, the feature loss enables our model (4) to generate visually better results that keep more final structures than our model (3), as shown in the last example in Figure~\ref{fig:real-mat}. This is consistent with many other works on image synthesis tasks that the feature loss tends to produce perceptually better results (often at the expense of the numerical performance)~\cite{blau2018perception, dosovitskiy2016generating, ledig2017photo, niklaus2018context,  sajjadi2017enhancenet,zhang2018unreasonable, zhu2016generative}.

When training our network with both the foreground decoder and the alpha decoder, color loss functions, namely $\mathcal{L}_{1}^{c}$ and $\mathcal{L}_{F}^{c}$, are naturally needed. By comparing models (4) and (6), (5) and (7) in Table~\ref{table:rst-com1k-matte}, we can find that these color losses can improve the alpha map estimation slightly. This is in part because the color and the alpha decoders share the same learned features, and the tasks of foreground color estimation and alpha map estimation are relevant.

\noindent\textbf{Data augmentation.} As shown in Table~\ref{table:rst-com1k-matte}, data augmentations, such as ReJPEGing and Gaussian blur, can greatly increases the errors of our methods on the Composition-1k testing dataset. On the other hand, we found that these data augmentations can greatly improve the generalization of our trained networks on the real world images. As shown in Figure \ref{fig:real-mat}, when trained with these data augmentation strategies, our models can maintain more fine details, such as hairs. Since these real world examples do not have ground truth, to obtain objective scores of these results, we evaluate the quality of composition results using our matting results. Specifically, we composite the foreground objects in source images onto some external background images using our matting results and then measure the visual quality of the composition results using the NIMA quality assessment algorithm~\cite{talebi2018nima}. As reported in Table~\ref{tbl:nima}, our data augmentation algorithms are helpful.  We also test our methods on the Spectral Matting dataset~\cite{levin2008spectral} with the known ground truth. This dataset is generated by photographing dolls in front of a computer monitor displaying seven different background images. The trimap is generated by dilating the alpha map by 20 pixels by alpha map denoising. Our method with DA outperforms our method without DA significantly according to most of the metrics: 3.58 vs 4.28 (SAD), 6.64 vs 9.05 (MSE), and 2.57 vs 3.19 (Conn), and slightly reduces the performance according to Grad: 2.04 vs 1.92.

\begin{figure*}[t]
	\centering
	\includegraphics[width=1.0\textwidth]{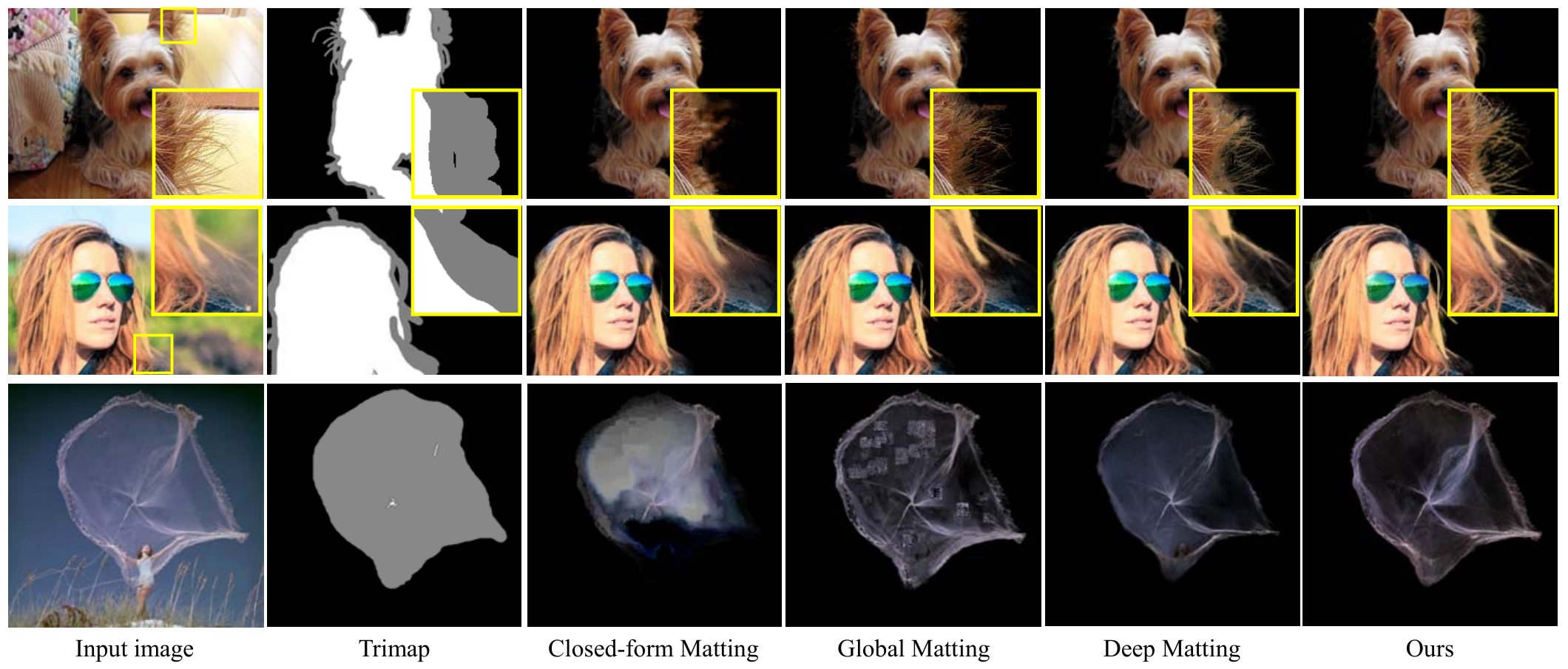}\vspace{-0.06in}
	\caption{Comparison of the composite results on the real world image dataset\cite{xu2017deep}.}\vspace{-0.2in}
	\label{fig:real-fg}
\end{figure*}

\subsection{User study}
\label{sec:study}

To further evaluate the quality of our results, we conducted a user study. We compared our method (model (7)) with three representative methods, including Deep Matting ~\cite{xu2017deep} and two state-of-the-art non-deep learning methods, Close-form Matting~\cite{levin2008closed} and  Global matting~\cite{he2011global}.

Our study used all the 31 real-world images from Xu~\etal\cite{xu2017deep}. We used a similar protocol to Xu~\etal\cite{xu2017deep} to produce the results for the study. For the methods except Deep Matting, we composite the predicted foreground and alpha map onto a blank background image. We use the black background or the white background randomly with the exceptions that for certain foreground images, a particular background color is not appropriate. For example, it is meaningless to composite the black hair onto a black background image, so for such an example, we choose to use the white background. Since Deep Matting does not output the foreground image, we composite the input image using the estimated alpha map as suggested in their paper. Therefore, the comparison between our results with those from Deep Matting should be interpreted with a grain of salt. 

Our user study recruited 42 students with different backgrounds. None of them have previous experience with the matting task. Therefore, we conducted a training session for each participant before the formal study. Specifically, each of them was shown two real-world images. For each image, we showed two matting results from different methods without revealing which methods were used to generate these results. We then explained the differences between two results to the participant. This training session is helpful as the subtle difference in matting results was often difficult to spot for people with no prior matting experience.

In our study, we divided the 42 participants into three groups. Each group evaluated how our results compared to one of the three existing methods. In each trial, a participant was presented with a screen that only shows a source image and two corresponding matting results at a time. The participant could select which image to view by clicking the corresponding button or using the \emph{left} or \emph{right} key on the keyboard. In this way, the participant can flip between different images to examine the quality or compare the difference. In each trial, the participant was asked to choose a more accurate and realistic result between the two results. Each participant conducted 31 trials so that the results for all the 31 testing images are evaluated.

We calculated the percentage of the times that our results were preferred by the participants and then calculated the average and the standard deviation for each group. As reported in Table~\ref{table:userstudy}, more of our results are preferred by the participants than all the comparing methods. Figure~\ref{fig:real-fg} shows some examples in our study. They show that our method can better capture very fine structures like the hair in the first example even when the hair shares a similar color to the background. In the last example, our result not only keeps the delicate edge of the lace, which is lost in the other results, but also is free from the color bleeding problem where the blue background color contaminated the result.

\begin{table}[t]
	\centering
	\caption{The user study in the real world image dataset~\cite{xu2017deep}.}
	\label{table:userstudy}
	\begin{tabular}{lcc}
			\hline
			 Ours vs &Mean preference rate & Std \\
			\hline

			Global Matting \cite{he2011global} & 85.48\%  &0.21  \\
			Closed-form Matting \cite{levin2008closed} & 84.11\% &0.19 \\
			Deep Matting \cite{xu2017deep} &77.67\%  &0.24 \\
			\hline
	\end{tabular}\vspace{-0.2in}
\end{table}

\section{Conclusion}

This paper presented a context-aware deep matting method for simultaneously estimating the foreground and the alpha map from a single natural image. We developed a two-encoder-two-decoder neural network for this task. The two encoders were designed to capture both the local fine structures and the more global context information to disambiguate the foreground and background with a similar appearance. The two decoders output the foreground and the alpha map respectively. Our experiments showed that using the feature loss helps to obtain visually more pleasant matting results while the Laplacian loss tends to optimize the numerical performance. Our experiments also showed that dedicated data augmentation methods, such as Re-JPEGING and Gaussian blurring, are helpful to generalize the neural network trained on a synthetic dataset to handle real-world challenging matting tasks.

\vspace{0.05in}
\noindent\textbf{Acknowledgments.}
The source images in Figure 1 are used under a Creative Commons license from Flickr users Robbie Sproule, MEGA PISTOLO and Jeff Latimer. The background images in Figure 1 are from the MS-COCO dataset~\cite{lin2014microsoft}. Source images used in Figure 2, 3, 4, 6, and 7 are from the matting dataset shared by Xu \etal\cite{xu2017deep}. We thank Nvidia for their GPU donation and Google for their cloud credits. 

{\small
\bibliographystyle{ieee_fullname}
\bibliography{egbib}
}

\end{document}